\documentclass[a4paper,conference]{IEEEtran_icpr}

\usepackage[pagebackref=true,breaklinks=true,letterpaper=true,colorlinks,bookmarks=false]{hyperref}
\pdfoutput=1

\ifCLASSOPTIONcompsoc
  \usepackage[nocompress]{cite}
\else
  \usepackage{cite}
\fi
\usepackage{times}
\usepackage{graphicx}
\usepackage{amsmath}
\usepackage{amssymb}
\usepackage{mathtools}
\usepackage[export]{adjustbox}
\usepackage{framed}
\usepackage{mdframed}

\usepackage{multirow}
\usepackage{bm}
\usepackage{amsbsy}

\usepackage{fixltx2e}
\usepackage[font=normal]{caption}  

\begin{document}
\def\SP#1{\textsuperscript{#1}}
\def\SB#1{\textsubscript{#1}}
\def\SPSB#1#2{\rlap{\textsuperscript{#1}}\textsubscript{#2}}

\providecommand{\e}[1]{\ensuremath{\times 10^{#1}}}

\title{To Boost or Not to Boost? On the Limits of Boosted Trees for Object Detection}


\author{\IEEEauthorblockN{Eshed Ohn-Bar and Mohan M. Trivedi}
\IEEEauthorblockA{Computer Vision and Robotics Research Laboratory\\
University of California San Diego\\
\{eohnbar, mtrivedi\}@ucsd.edu}}

\markboth{}%
{Shell \MakeLowercase{\textit{et al.}}: Bare Demo of IEEEtran.cls for Computer Society Journals}

\maketitle

\begin{abstract}
We aim to study the modeling limitations of the commonly employed boosted decision trees classifier. Inspired by the success of large, data-hungry visual recognition models (e.g. deep convolutional neural networks), this paper focuses on the relationship between modeling capacity of the weak learners, dataset size, and dataset properties. A set of novel experiments on the Caltech Pedestrian Detection benchmark results in the best known performance among non-CNN techniques while operating at fast run-time speed. Furthermore, the performance is on par with deep architectures (9.71\% log-average miss rate), while using only HOG+LUV channels as features. The conclusions from this study are shown to generalize over different object detection domains as demonstrated on the FDDB face detection benchmark (93.37\% accuracy). Despite the impressive performance, this study reveals the limited modeling capacity of the common boosted trees model, motivating a need for architectural changes in order to compete with multi-level and very deep architectures. 
\end{abstract}

\IEEEpeerreviewmaketitle

\section{Introduction}
\label{sec:intro}

\begin{figure}[!t]
\centering
\begin{tabular}{c}
\includegraphics[trim =0mm 95mm 11mm 0mm, clip=true,width=3.4in]{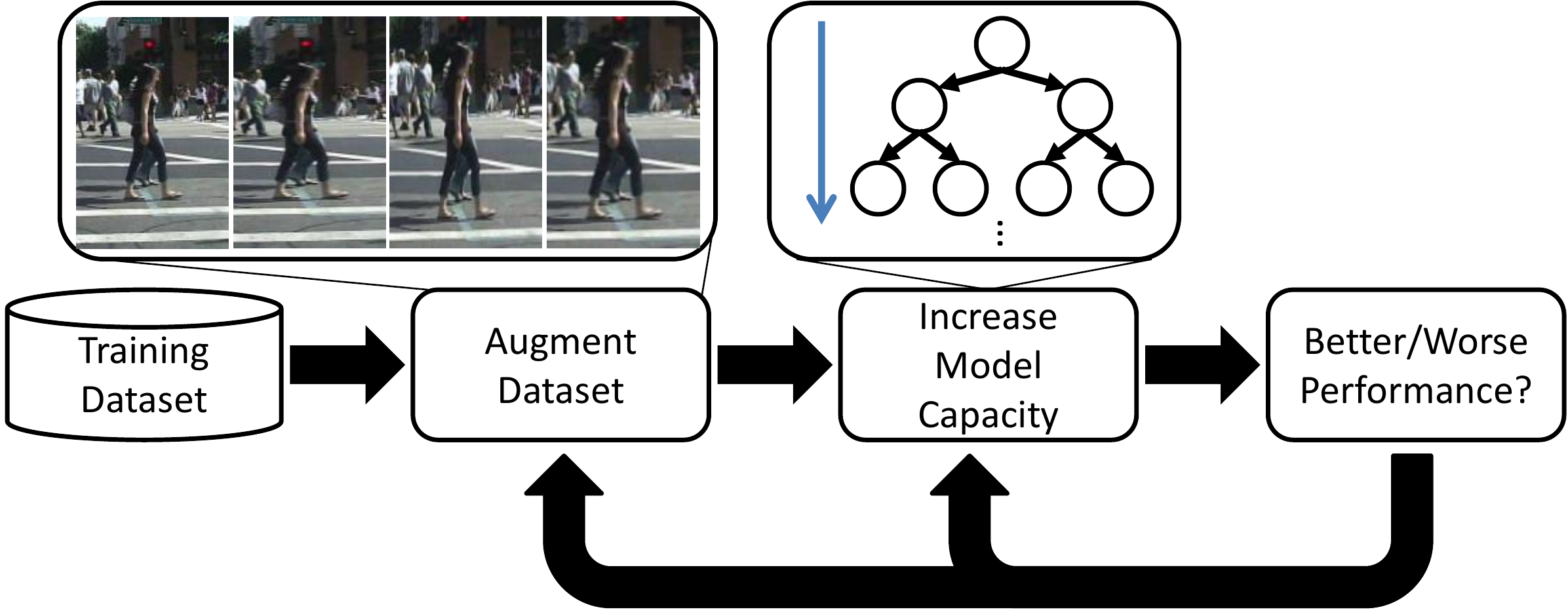}
   \end{tabular}
 \caption{By varying dataset properties with different augmentation techniques (a scaling example is shown) and model capacity (tree depth), we study limitations of the boosted decision tree classifier. Consequently, the insights revealed are employed in order to train state-of-the-art pedestrian and face detectors.}
 \label{fig:fig1}
\end{figure}

The increase in data availability and computational power have led to a revolution in object recognition. Modern object recognition algorithms combine deep multi-layer architectures with large datasets in order to achieve state-of-the-art recognition performance \cite{imagenet,sermanet13,simonyan2014very}. A main advantage of such learning architectures stems from the architectural ability to increase modeling capacity (e.g. network parameters and depth) in a straightforward manner. At the same time, boosted detectors \cite{vj,6977320,dollar2012pedestrian} remain highly successful for fast detection of objects \cite{zhang2015efficient,de2014combinator,tenyerspeds,gpuicf,zhang2014center,jones2008pedestrian,icf,acostea,DollarPAMI14pyramids,ldcf,vehicles_TITS15,spatialpooleecv,ccf,compACT,costea2015fast}.
The Viola and Jones \cite{vj} learning architecture has remained largely unchanged, with boosting \cite{boost} used for training a cascade of weak learners (e.g. decision trees). State-of-the-art pedestrian detectors often employ such models \cite{DollarPAMI14pyramids}, while pursuing improved feature representations in order to achieve detection performance gains \cite{tenyerspeds,ldcf}. Despite being a different learning architecture compared to the modern Convolutional Network (CNN) \cite{imagenet}, the boosted decision tree model also allows for a straightforward increase in modeling capacity for handling large datasets (Fig. \ref{fig:fig1}). Motivated by this observation, we perform an extensive set of experiments designed to better understand the limitations of the commonly employed boosted decision trees model. The large Caltech pedestrian benchmark \cite{dollar2012pedestrian} is suitable for such a task, with some preliminary work \cite{ldcf,checkerboard} showing significant performance gains by increasing dataset size and modeling capacity of the weak learners. 

\begin{figure}[!t]
\centering
\begin{tabular}{c}
\includegraphics[width=3.4in]{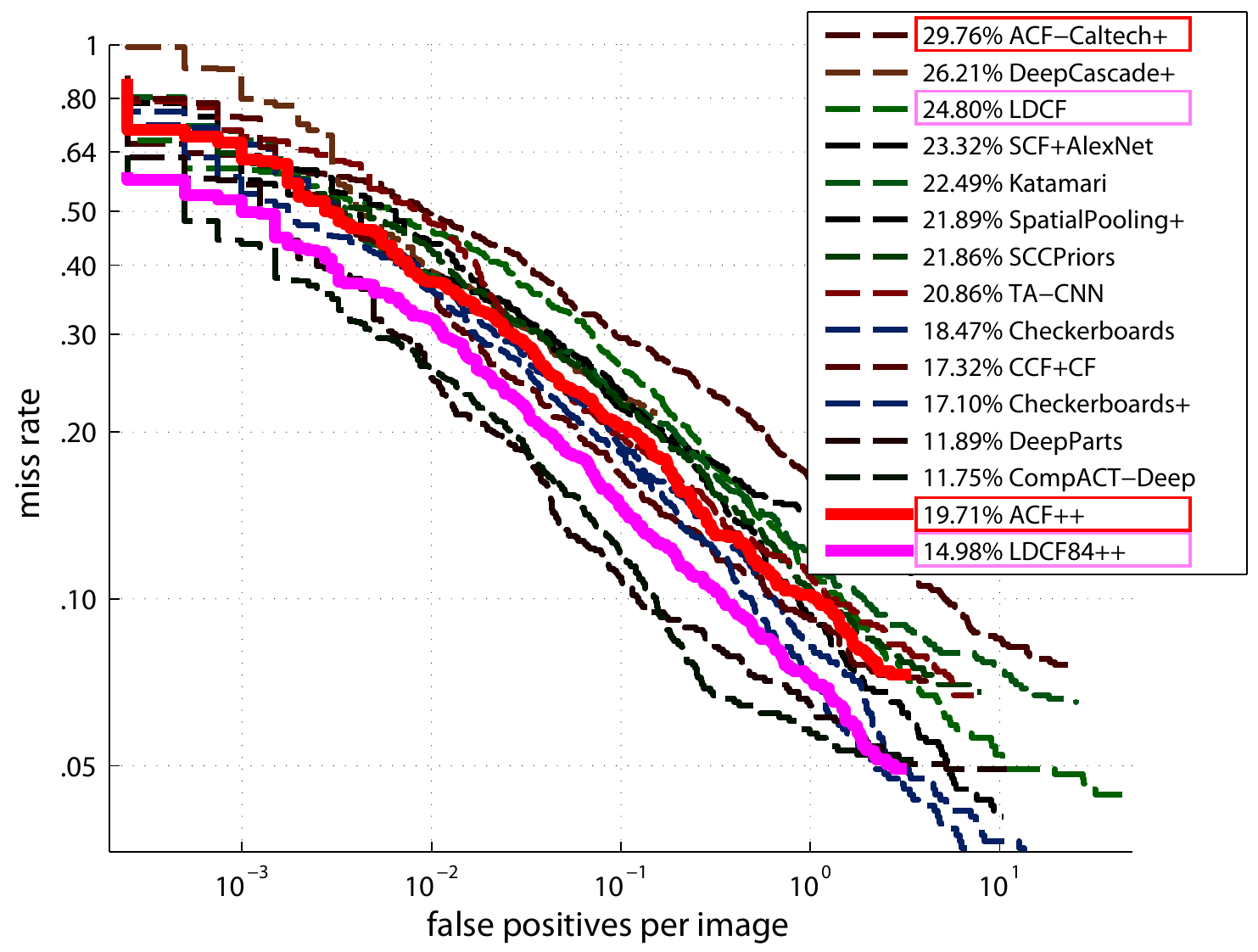}
   \end{tabular}
 \caption{Comparison of our key results (in thick lines, ACF++ and LDCF84++) with published methods on the Caltech test set. Our version of the popular ACF and LDCF models demonstrate large gains in detection performance over the baselines with no significant modifications to the model or feature extraction pipeline.}
 \label{fig:usareason}
\end{figure}

 In particular, the contributions of this work are as follows.

\textbf{Limitations of a classifier}: The relationship between increasing the modeling capacity of the decision trees jointly with dataset size has been established in previous literature \cite{boost,ldcf,Das_ITSC14}, yet its limits have not been fully explored. Surveying current literature, we found inconsistent or limited usage of this relationship in boosting for object detection. In particular, it is not known whether the modeling capacity has saturated or whether it can be further increased with additional data for the detection task. Therefore, we sought to perform a rigorous study with the relationship between dataset size and model capacity as the main research question. In the process of studying this problem, insights regarding the limitations of the model arise. The conclusion of this study is mostly negative, as increasing the modeling capacity of the weak learners in the current boosted decision tree detector provides limited generalization power. Saturation of performance occurs quickly, with no further gains shown with additional data augmentation. Therefore, architectural changes are needed in order to resolve the issue of efficiently increasing modeling capacity in the boosted tree classifier. 




\textbf{Performance gains}: The experiments in this paper result in consistent gains in object detection performance, insights into optimal dataset properties, and best training practices on Caltech. For instance, we demonstrate that the commonly employed video-based augmentation \cite{ldcf,checkerboard} on the Caltech training set is limited when compared to other augmentation options. With no significant modifications to the learning or feature extraction pipelines, we improve over the Aggregate Channel Feature (ACF) detector \cite{DollarPAMI14pyramids} baseline by $\sim$10\% points while still employing the 10 HOG+LUV feature channels. A simple filtering of the channels with 4 filter banks (LDCF \cite{ldcf}) results in the best performing non-CNN detector on Caltech to date. The results of the study are shown to generalize to face detection as well. Note that we intentionally avoid using deep CNN features here, as these often involve training on a large external data. On the other hand, simple channel features \cite{DollarPAMI14pyramids} allow for isolation of the strengths, limitations, and the role of the dataset on the boosting classifier. Furthermore, the careful examination of data augmentation and model capacity allows for a more appropriate comparison between CNN (where extensive data augmentation is commonly employed) and non-CNN detectors on datasets such as Caltech. We hope the more meaningful comparison will motivate future developments in object detection. We also note that several recent CNN-based models on Caltech \cite{hosang2015taking,rotfilt,deepparts} make use of the boosted trees classifier.   


\section{Experimental Settings}

\subsection{Boosting Framework}
\label{subsec:boost}

This section presents the tools that will be used for the remainder of the study. Boosting \cite{boost} involves greedily minimizing a loss function for the following classification rule

\begin{equation}
	\mathbf{H}(\mathbf{x}) = \sum_t \alpha_t \mathbf{h_t}(\mathbf{x})
\end{equation}

where $\mathbf{h_t}(\mathbf{x})$ are weak learners, $\alpha_t$ are scalars, and $\mathbf{x} \in \mathbb{R}^K$ is a feature vector. In this work, we employ the commonly used soft cascade with decision trees as the weak learners \cite{bourdev2005robust,bagging2}. These are composed of decision stumps such that each non-leaf node $j$ produces a binary decision using a feature index $k$, a threshold $\tau \in \mathbb{R}$, and a polarity $p \in \{\pm 1 \}$, $h_j(\mathbf{x}) \equiv p_j sign(\mathbf{x}[k_j] - \tau_j)$. An important parameter is the maximum tree depth, which was generally taken to be 2 for object detection, yet recent studies \cite{ldcf,checkerboard,spatialpool} propose depths of 3-4 as suitable to accommodate an increase in dataset size. Hence, our first issue is efficient training of boosted models on large datasets. The need for quick training is key to algorithmic development and competitive performance, especially when considering current state-of-the-art CNN object detections are trained on millions of samples.


\textbf{Training with randomness}: In this work, we adopt the quick boosting approach from \cite{quickboost}. Because stump training is costly, $\mathcal{O}(K \times N)$ for $K$ features and $N$ samples, a common practice is to quantize feature values into bins (256 in this work) and sample a random subset of the features when searching for the optimal feature index. Most approaches employ the ACF detector \cite{DollarPAMI14pyramids}, which samples 1/16 of the total feature set in each iteration. We observed significant variation in performance over multiple runs with this random sampling strategy, by up to 3-4\%. This is an issue, yet an appropriate discussion was not found in related literature. Therefore, we modify the last stage out of four of generating hard negatives with increasing number of weak classifiers, $[ 64, 256, 1024, 4096]$. In the last stage, the feature set is exhaustively searched over for the \textit{initial 512 weak learners}, after which random sampling of 1/16 is followed. This procedure was found to be necessary for careful analysis, and it balances reproducibility with speed (unlike the very slow exhaustive search in all iterations \cite{checkerboard}) reducing variability to within $\sim$1\% in performance. In experiments where significance is uncertain, multiple random seeds are used for training initialization, with the best one shown. Batch training is another possible direction which we leave for future work.


\subsection{Augmentation Techniques}
\label{subsec:augment}

As previously mentioned in Section \ref{sec:intro}, large gains in boosting-based pedestrian detection came from dataset size increase and increased depth of the decision tree (introduced in Nam \textit{et al.} \cite{ldcf}), although the design choices for these have not been well justified or studied. Consequently, existing state-of-the-art detectors employ dense frame sampling (e.g. sampling every $3^{rd}$ frame resulting in a ten-fold increase in dataset size, Caltech$\times10$), yet this type of video-based augmentation will be shown to present several issues leading to sub-optimal detector performance. 


Identifying the most successful augmentation techniques is essential, as the increased dataset size regularizes training of models with increased capacity. We note that others have studied data augmentation, specifically for pedestrian detection, but mostly with a goal of lowering annotation effort (i.e. render pedestrian images into real backgrounds \cite{6977491,vazquez2012unsupervised}) and not improved performance.


In addition to sampling more frames from the Caltech videos, the following augmentations are studied:

\textbf{Color}: Color alteration has been proposed in \cite{imagenet} using Principal Component Analysis (PCA) on the RGB pixel values in the training set. Specifically, a random Gaussian variable, $\alpha \in \mathbb{R}^3$ is drawn and the quantity $[\mathbf{p_1},\mathbf{p_2},\mathbf{p_3}] [\alpha_1 \lambda_1,\alpha_2\lambda_2,\alpha_3\lambda_3]$, where $\mathbf{p_i}$ and $\lambda_i$ are eigenvector-eigenvalue pairs, is added to the RGB image pixels.  

\textbf{Flipping}: Most current trained detectors employ horizontal mirroring of images in order to double the number of positive samples.  


\textbf{Translation (T)}: Two main parameters, namely the maximum amount of pixel shift allowed ($m$) and the sampling intervals ($n$), determine how many additional positive samples are generated. We note these in the experiments as T$nm$. For instance, T31 generates 4 additional samples at 1 pixel shift in each direction (west/east/north/south). 

\textbf{Scale+Crop (S)}: As experiments began, we discovered high sensitivity of the classifier to even small noise injected in the ground truth location. As an alternative to translation, we propose to scale each positive sample by a factor (either in the horizontal, vertical, or both directions), re-center, and crop it (shown in Fig. \ref{fig:fig1}). Hence, the procedure is slightly different from the common cropping employed in training CNN detectors (corner/center cropping). This augmentation is similar to slight ground truth width/height jitter without changes to the center of the box.


\textbf{Occlusion}: Occlusion handling is an important challenge for pedestrian detection \cite{whyandwhen,ohnbar_pedspatterns15}. Incorporation of samples with higher levels of occlusion is another technique for increasing dataset size and studying generalization capability. 

\textbf{FG/BG Transfer}: The visibility masks of pedestrians in Caltech can be employed in order to transfer foreground/background between images in the dataset. 


\subsection{Model Settings}

Our training settings specify a model size of 41 $\times$ 100 pixels. With padding, the sliding window becomes 64 $\times$ 128 pixels. In order to deal with smaller pedestrians (`reasonable' test settings involve pedestrians of height 50 pixels and up), we upsample the images by one octave. Most of the augmentation experiments will be performed on Caltech$\times3$ (sampling every $10^{th}$ frame from video), but higher samplings of Caltech$\times7.5$ (every $4^{th}$ frame), Caltech$\times15$, and Caltech$\times30$ (every frame) sampling will also be studied.

\section{Experimental Analysis}
\label{sec:expeval}

This section demonstrates the importance of data augmentation, impact of augmentation type, and significance of increased weak classifier model capacity in training a generalizable pedestrian detector. All results are measured in log-average miss rate (a lower value corresponds to better detection performance) as described in \cite{dollar2012pedestrian}.

\begin{figure*}[!t]
\centering
\begin{tabular}{ccc}
\includegraphics[width=2in]{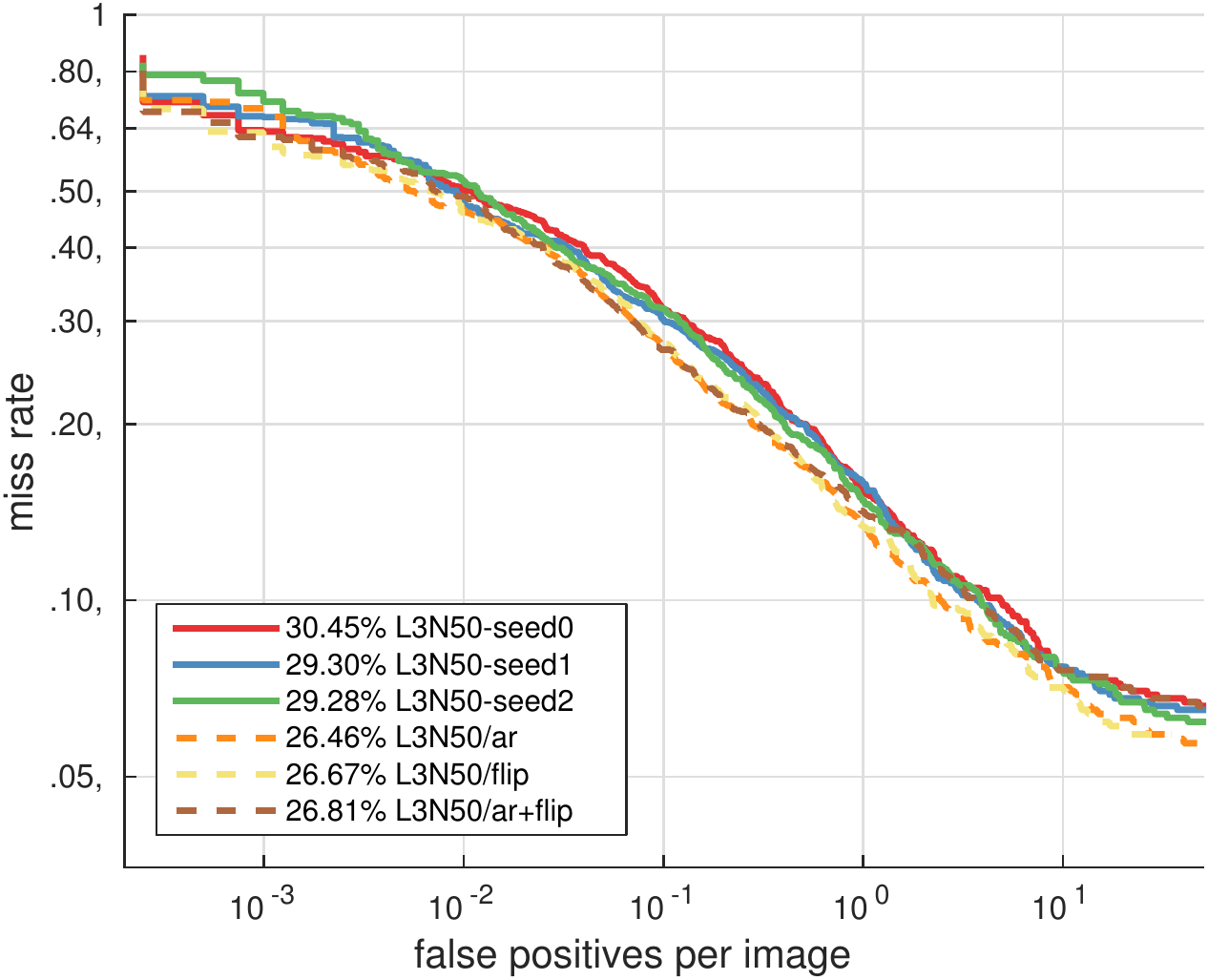}
&
\includegraphics[width=2in]{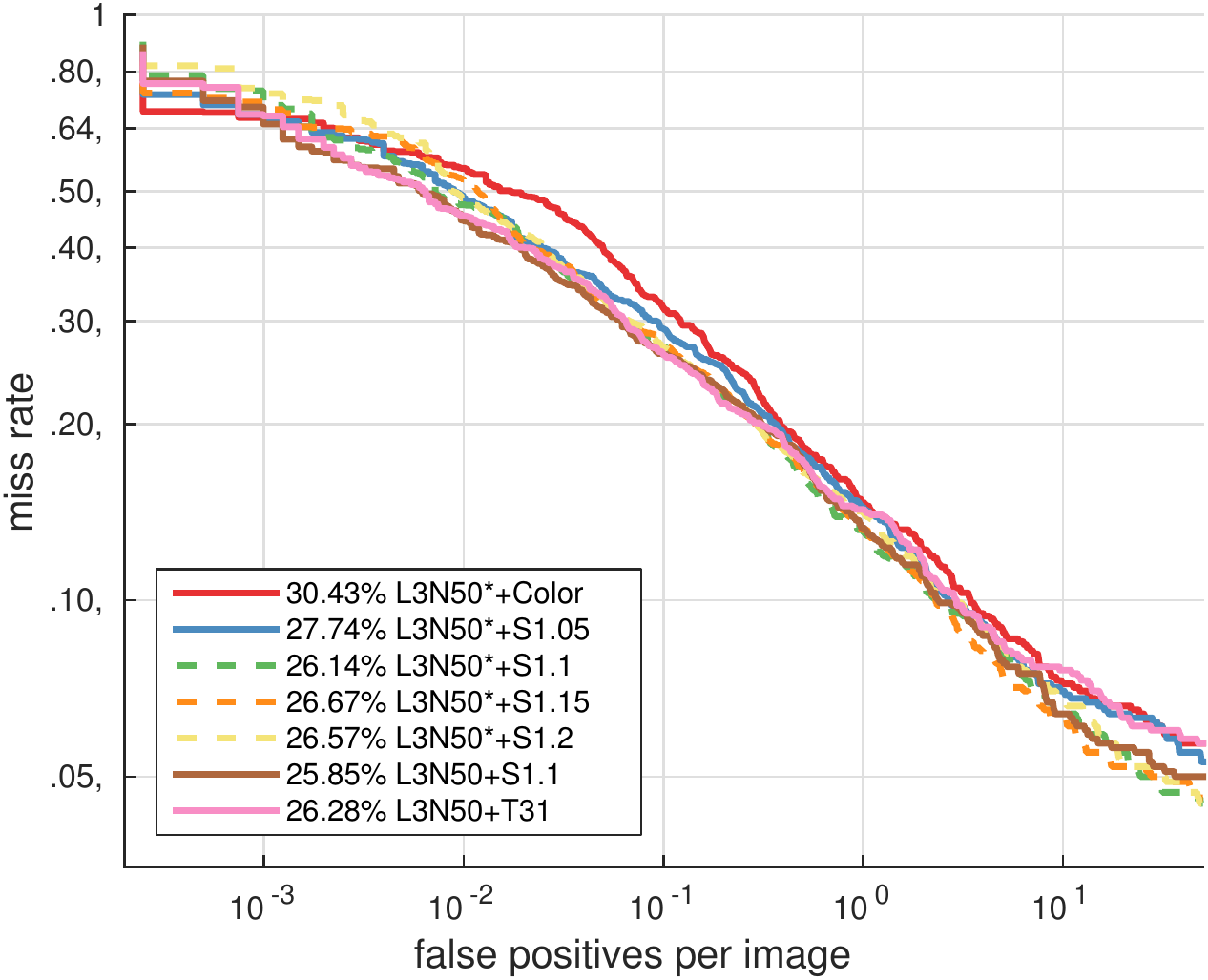}
&
\includegraphics[width=2in]{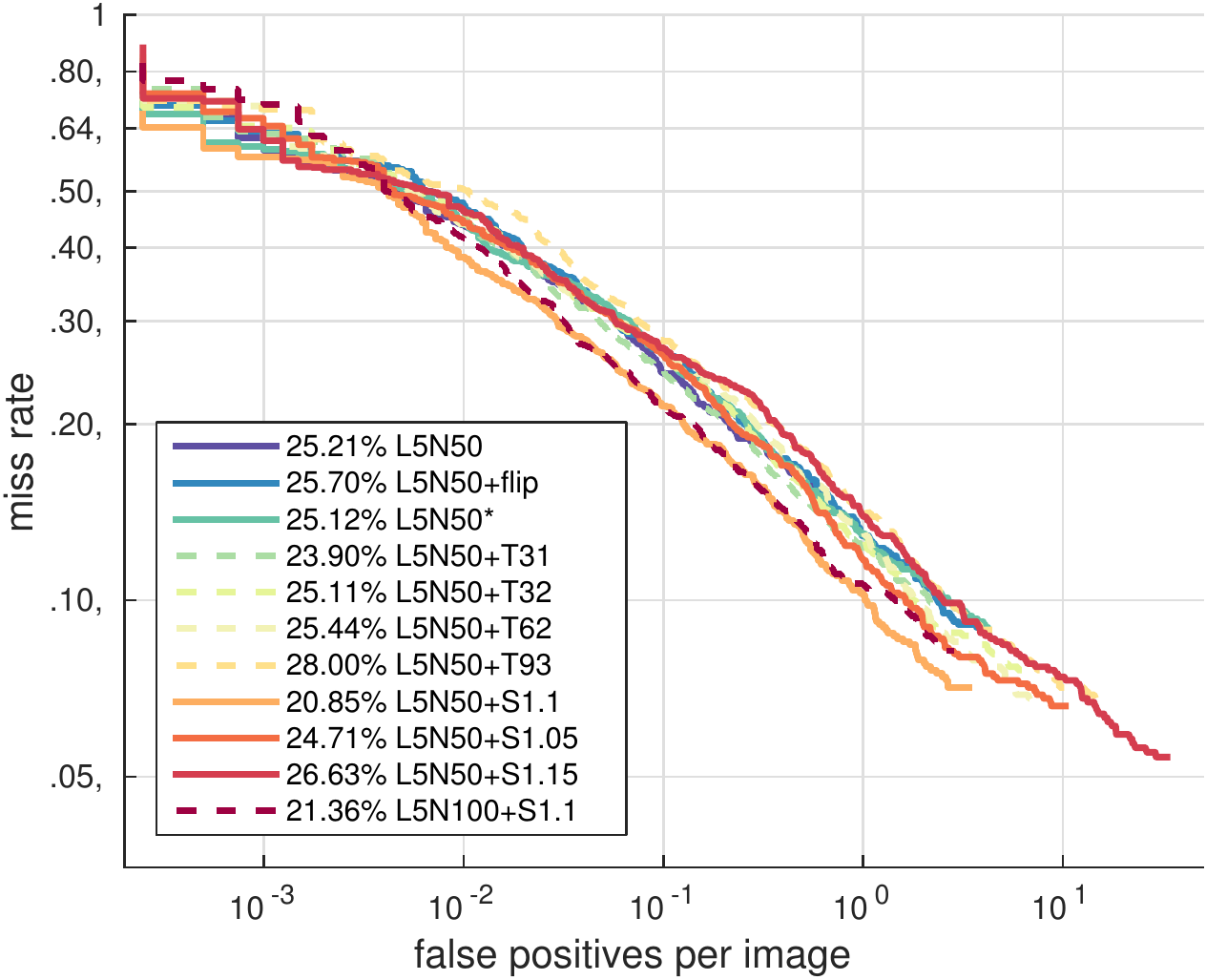}
\\
(a) & (b) & (c)
 \end{tabular}
 \caption{Experimental analysis of different training settings on Caltech pedestrians using Caltech$\times3$. }
 \label{fig:finalcurves}
\end{figure*}

\textbf{Baseline and notation}: Our experiments begin with Fig. \ref{fig:finalcurves}(a), where we first verify the impact of randomness with multiple random seeds in training (see Section \ref{subsec:boost}) on performance variation using Caltech$\times3$. The $L3N50$ model is the commonly trained maximum depth 3 model using the available implementation of \cite{dollar2012pedestrian} with 50k total negative samples (in each hard negative mining round we collect N/2 samples and replace these hard samples so that no more than N samples are trained over in a given round). We note that the only form of augmentation from Section \ref{subsec:augment} commonly employed in training boosting-based pedestrian detectors besides higher frame sampling from video is flip augmentation. Hence, flip augmentaiton is included in the $L3N50$ baseline. It achieves comparable results to the best known ACF results on Caltech of 29.76\% log-average miss rate (Fig. \ref{fig:usareason}).

\begin{figure*}[!t]
\centering
\begin{tabular}{ccc}
\includegraphics[width=2in]{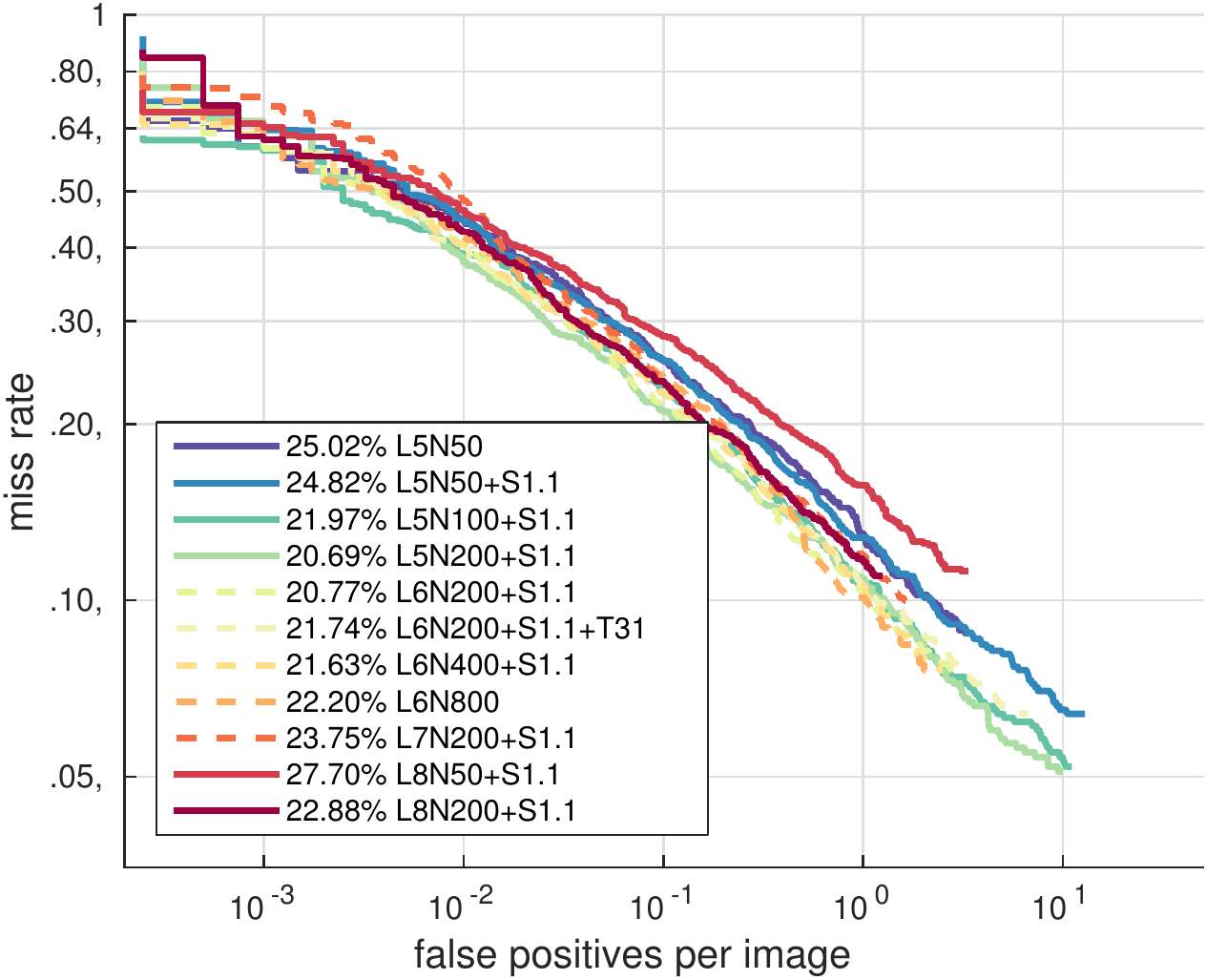}
&
\includegraphics[width=2in]{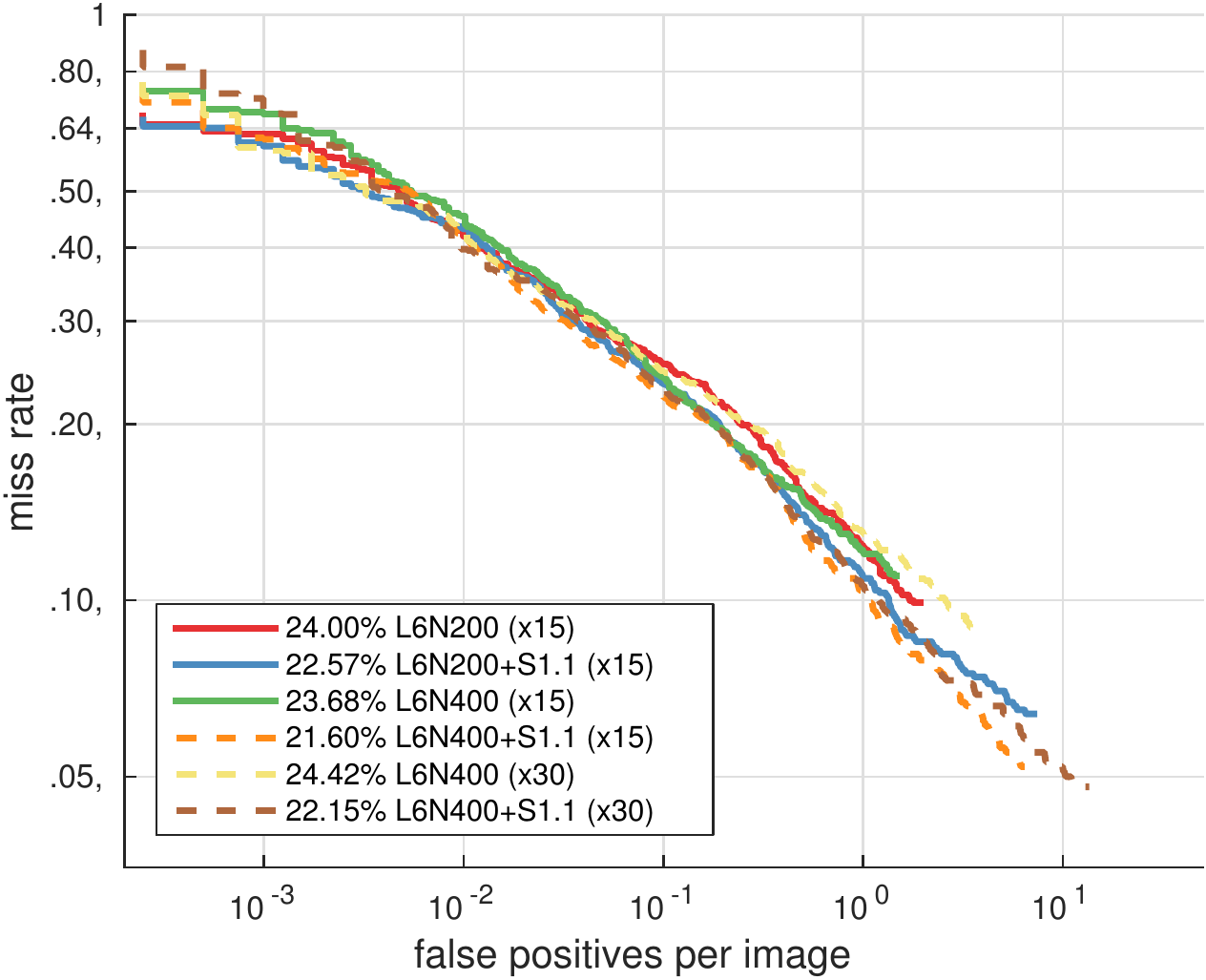}
&
\includegraphics[width=2in]{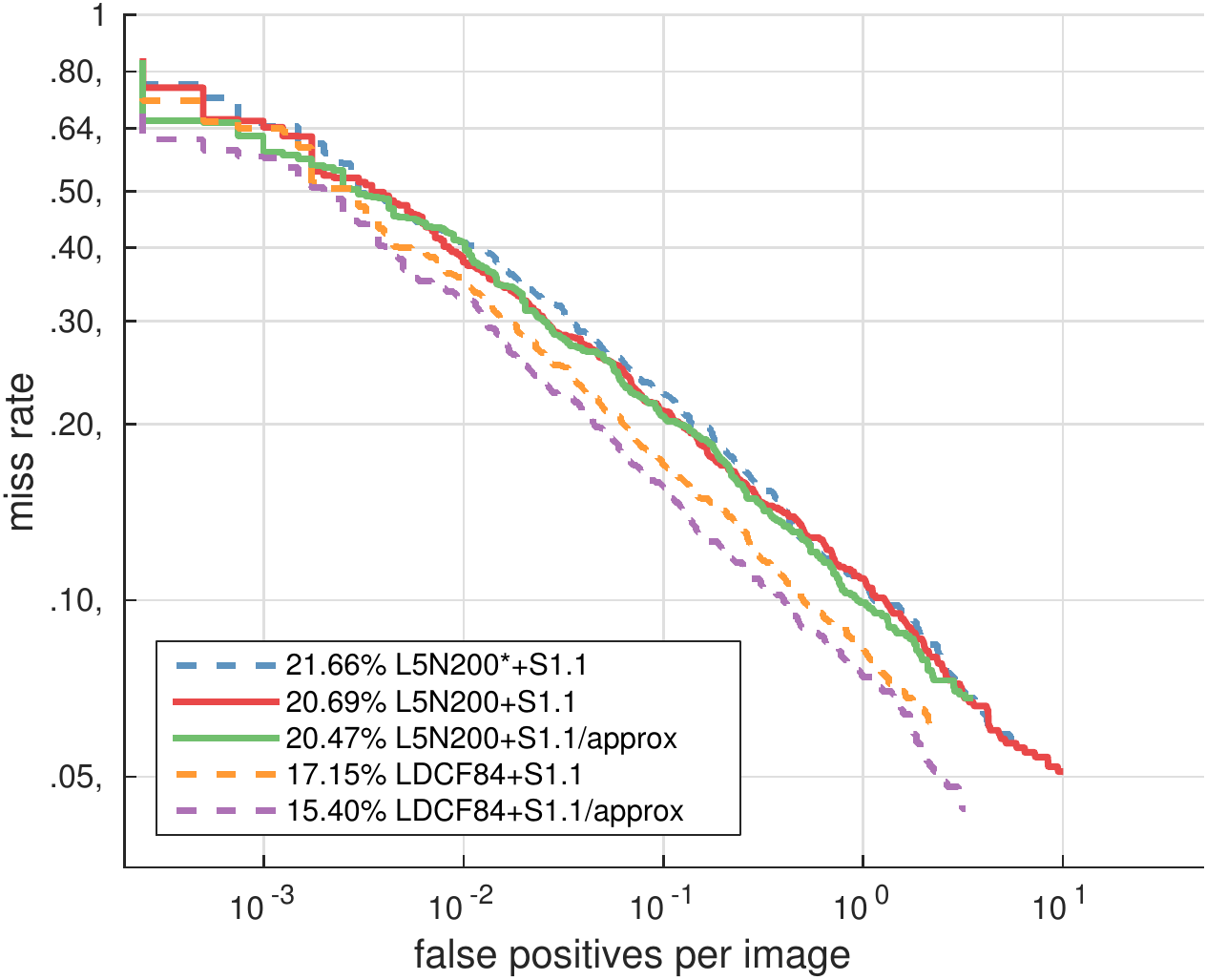}
\\
(a) & (b) & (c)
   \end{tabular}
 \caption{Experimental analysis of different training settings on Caltech pedestrians using additional video augmentation (Caltech$\times7.5$ unless stated).}
 \label{fig:finalcurves_caltech10}
\end{figure*}

 
\textbf{Flip augmentation}: Surprisingly, Fig. \ref{fig:finalcurves}(a) depicts how the removal of the flip augmentation in the $L3N50$ baseline results in a significant reduction in log-average miss rate from 29.28\% to 26.67\%! The reason for the performance reduction becomes apparent immediately, when aspect ratio (ar) standardization (used in training/testing all current object detectors on Caltech \cite{dollar2012pedestrian}) is removed during training in $L3N50/ar$. Specifically, the standardization is useful in training since it is enforced in evaluation, yet it leads to non-aligned bounding boxes. Flipping these boxes introduces additional poorly-aligned samples which the boosting model does not handle well, even with increased modeling capacity in later stages of the experiments (depth 5 or more models). The solution is simple, either train $L3N50/ar$ models (without aspect ratio standardization and with flip augmentation) or remove flipped sampling ($L3N50/flip$). Since this is a one-time initial modification, we slightly rename our `new baselines', so that $L3N50$ refers to $L3N50/flip$ (without flip, with aspect ratio standardization of training samples) and $L3N50*$ refers to $L3N50/ar$ (with flip, no aspect standardization). The two are shown to perform similarly in Figs. \ref{fig:finalcurves}(a) and \ref{fig:finalcurves}(b) (yet $L3N50/flip$ is faster to train as it contains half the samples). We note that this observation was not made in any of the related research studies, and consequently not employed by top performing boosted detectors.  

\textbf{Additional augmentation}: Depicted in Fig. \ref{fig:finalcurves}(b), the $L3N50$ and the $L3N50*$ models do not benefit from color augmentation, but do benefit from the scaling+crop scheme (best with a scaling factor of $S1.1$ in horizontal, vertical, and both directions) and translation augmentation (by 1 pixel shifts, $T31$). These experiments were run multiple times with different random seeds to ensure statistical significance. The improvements will be shown to be more dramatic with increased modeling capacity (deeper trees). We make two final notes, increasing the number of negatives to $L3N100$ does not further improve performance, and overall performance has reached a plateau. Therefore, the modeling capacity is increased in $L5N50$. 

\textbf{Increased model capacity}: Still on Caltech$\times~3$, the analysis of Fig. \ref{fig:finalcurves}(c) demonstrates a consistent trend with the previous experiments. Note how flip augmentation in $L5N50+flip$ is still not beneficial, but it is much better handled. The best results are reported in Fig. \ref{fig:finalcurves}(c). We observe a significant jump in performance reaching 20.85\% with the fast ACF model. These are the best reported results to date (nearly optimal from our experiments). Scaling is shown to be more useful than translation, as even slight off-center shifts hinder performance and are not handled well by the classifier. This is encouraging, yet we begin to understand how limited the boosting model is in sensitivity to dataset properties. Samples with even minimal occlusion and FG/BG transfer were also not shown to be beneficial for increasing the dataset size. Furthermore, we experimented with addition of external pedestrian datasets (e.g. ETH \cite{eth}) and increasing tree depth, but no benefit was shown.  

  \textbf{Video augmentation}: A main conclusion is that very little video augmentation (Caltech$\times3$ vs. Caltech$\times10$ in current state-of-the-art detectors \cite{checkerboard,ccf,ldcf}) was needed to reach best performance. Denser sampling of every $4^{th}$ frame is shown in Fig. \ref{fig:finalcurves_caltech10}(a), demonstrating an interesting trend. First, we must increase the number of negatives to $200k$ ($N200$) to reach comparable performance. This fact exposes another known limitation of currently used boosting models which is not reported in related research, one of dataset imbalance. This is necessary for implicitly regularizing deeper models (shown up to depth 8 in Fig. \ref{fig:finalcurves_caltech10}(a)). Second, we observe that the $S1.1$ scheme provides consistent improvements across different video sampling strategies and model choices (Fig. \ref{fig:finalcurves_caltech10}(b)). This demonstrates the sub-optimal yet commonly employed procedure of video augmentation. Third, additional augmentation or increased model capacity does not improve over the 20.69\% of $L5N200$. To ensure no further gains can be made, we further increase tree depth to 6 to handle additional data and further increase dataset in Fig. \ref{fig:finalcurves_caltech10}(b), with no visible improvement. Increasing tree depth beyond 6 leads to overfitting behavior, and additional data must be sampled (although our experiments show no final gains).
  
  \textbf{Better features}: The achieved 20.69\% miss rate is the best known ACF results to date, nearly matching the more computationally intensive feature-rich Checkerboards (CB) detector \cite{zhang2015filtered} (18.47\% miss rate). CB employs 61 filters \textit{for each} of the 10 core HOG+LUV channels, resulting in significant increase to computational requirements. To further study the proposed modifications, we extract features using 4 $8 \times 8$ LDCF filters \cite{ldcf} computed using PCA eigenvectors of feature patches (referred to as the \textbf{LDCF84} model). The method still runs 5 times faster than the CB comparison, while reaching a 17.15\% miss rate. This demonstrates the generalization of the proposed best practices to other, feature-rich approaches as well. As a final experiment, we report results without the approximation of features in the multi-scale feature pyramid \cite{DollarPAMI14pyramids}. This allows for a more clear performance comparison against other methods \cite{rotfilt,zhang2015filtered} which do not employ approximation. The gains are even more impressive, reaching 15.40\% with a light-weight LDCF84 model, significantly outperforming CB.

  \textbf{Summary}: Despite impressive gains in performance stemming from simple, intuitive, and theory-driven considerations in training, the boosted detector can only handle mild deformations as augmentation. For instance, color jitter is not helpful, flipping is problematic (and removed in most of the experiments without reduction in performance), and slight translation shifts produce a more difficult learning task that is not modeled well even with deeper trees. In order to address these limitations, we proposed to use an augmentation technique which maintains centered ground truth boxes. Dataset size is key in training deeper trees, yet performance saturates at depth 5/6 even with extensive augmentation. Hence, unlike additional layers in CNN architectures, further increase of the tree depth provides a limited benefit in terms of classification power. The experiments motivate future work for the effective increase of the modeling capacity of tree models. 
  
  \textbf{Run-time analysis}: Although not the main focus of the study, we report run-time for the interested reader. The ACF-$L5$ and LDCF84 models run on a CPU at 6.7 and 2.5 frames per second (fps), respectively. These models are significantly faster than all of the top performing feature-rich models, which range in 0.1-1 fps (e.g. CB \cite{checkerboard} runs at 0.5 fps).
  


\subsection{Context Analysis}

Context in the form of scene representation is essential for robust pedestrian detection. The deeper tree models are able to better capture relationships between features in the image. In particular, we have observed how deeper models are more likely to select features in the padding area of the model around the pedestrian. As a final limitation experiment on Caltech, we propose to use a location prior model in order to study to limitations of the model in capturing spatial context. 

First, we studied increasing model padding further with tree depth as means of better representing contextual information, yet this did not impact performance. On the other hand, due to the application domain, it is reasonable to study context in terms of perspective and position of objects. A second approach was shown to improve performance further, hinting that there is still room for future improvements in context modeling.

Given a detection, $\mathbf{p} = \begin{bmatrix} x& y& w& h& s\end{bmatrix}^\intercal$, we construct $\mathbf{\phi_p} = \mathbf{p}\mathbf{p}^\intercal$ for each training sample. Consequently, entries in $\mathbf{\phi_p}$ are employed as features for capturing relationship between position, size, confidence features, and their products. The object score, $s$, is recomputed using an SVM \cite{svm} on $\mathbf{\phi_p}$. We note that this generalizes (and outperforms) the approach in \cite{multires}, which only employs a subset of the proposed $\mathbf{\phi_p}$ feature set (setting $\mathbf{\phi_p} = \begin{bmatrix} h^2 & y^2& hy& h& y \end{bmatrix}$).


As shown in Fig. \ref{fig:usareason}, context integration consistently improve performance of both the ACF and LDCF84 models by a further 0.5-1\% reduction in miss rate. Hence, there is still room for improvement in order to better capture contextual cues within the models. We term our models, ACF++ and LDCF84++, where the first `+' stands for the final models trained with data augmentation (Fig. \ref{fig:finalcurves_caltech10}), and the second `+' for the contextual reasoning. We observe how the proposed modifications produce consistent improvements over both the ACF and LDCF baselines in Fig. \ref{fig:usareason} by about 10\%. The proposed high-performing models are new baselines in a way, and further gains may be obtained with combined with insights from other studies listed in Fig. \ref{fig:usareason}. 

\subsection{Results on Caltech-New}

The proposed techniques are tested on the improved annotation Caltech pedestrians dataset presented in \cite{rotfilt}, as shown in Table \ref{tab:caltech}. The cleaner annotations have a great impact on training the ACF++ and LDCF++ models, which is consistent with our observations in previous experiments. Consequently, the best detection results to date are achieved among both CNN and non-CNN methods on the challenging MR\SPSB{N}{-4} metric, with a miss rate of 18.29\%.  


\begin{table}[t!]
\centering
\caption{Results on the new annotations on Caltech \cite{rotfilt} using log average miss rate over $[10^{-2},10^0]$ and $[10^{-4},10^0]$ (MR\SPSB{N}{-2} and  MR\SPSB{N}{-4}, respectively). Our ACF+ (with data augmentation) and ACF++ (with contextual reasoning) is on par with state-of-the-art detectors while enjoying low computational complexity. Our LDCF84++ employs just 4 filters on top of the ACF features and outperforms the VGG baseline by a large margin on the more challenging metric MR\SPSB{N}{-4}. }
\label{tab:caltech}
\begin{tabular}{|l c|} 
 \hline
 
 Method & MR\SPSB{N}{-2} ( MR\SPSB{N}{-4} ) \\ 
 \hline\hline
 RotatedFilters \cite{rotfilt} & 12.87 (24.10) \\ 
 RotatedFilters+VGG \cite{rotfilt} & \textbf{9.32} (21.72)  \\
 ACF+  & 15.17 (27.28) \\
 ACF++  & 13.27 (25.26)  \\
 LDCF84+ & 11.76 (21.08) \\ 
 LDCF84++ & {9.71 (\textbf{18.29})} \\ 
 \hline
\end{tabular}
\end{table}

\begin{table}[t!]
\centering
\caption{Improvement due to incorporation of the insights in this paper when training face detection models.}
\label{tab:fddb}
\begin{tabular}{|l c|} 
 \hline
 \multicolumn{2}{|c|}{Results on WIDER-validation}
 \\
 \hline
 Method & AP (\%) \\ 
 \hline
 ACF-$L3N50*$  & 27.97  \\
 ACF-$L9N200*$  & 34.64 \\
 ACF-$L9N200*$+S1.1 (ACF+)  & 46.80 \\
 LDCF84-$L9N200*$+S1.1 (LDCF+)  & \textbf{47.66} \\
 \hline
\end{tabular}
\end{table}

\begin{figure*}[!t]
\centering
\begin{tabular}{ccc}
\includegraphics[width=2in]{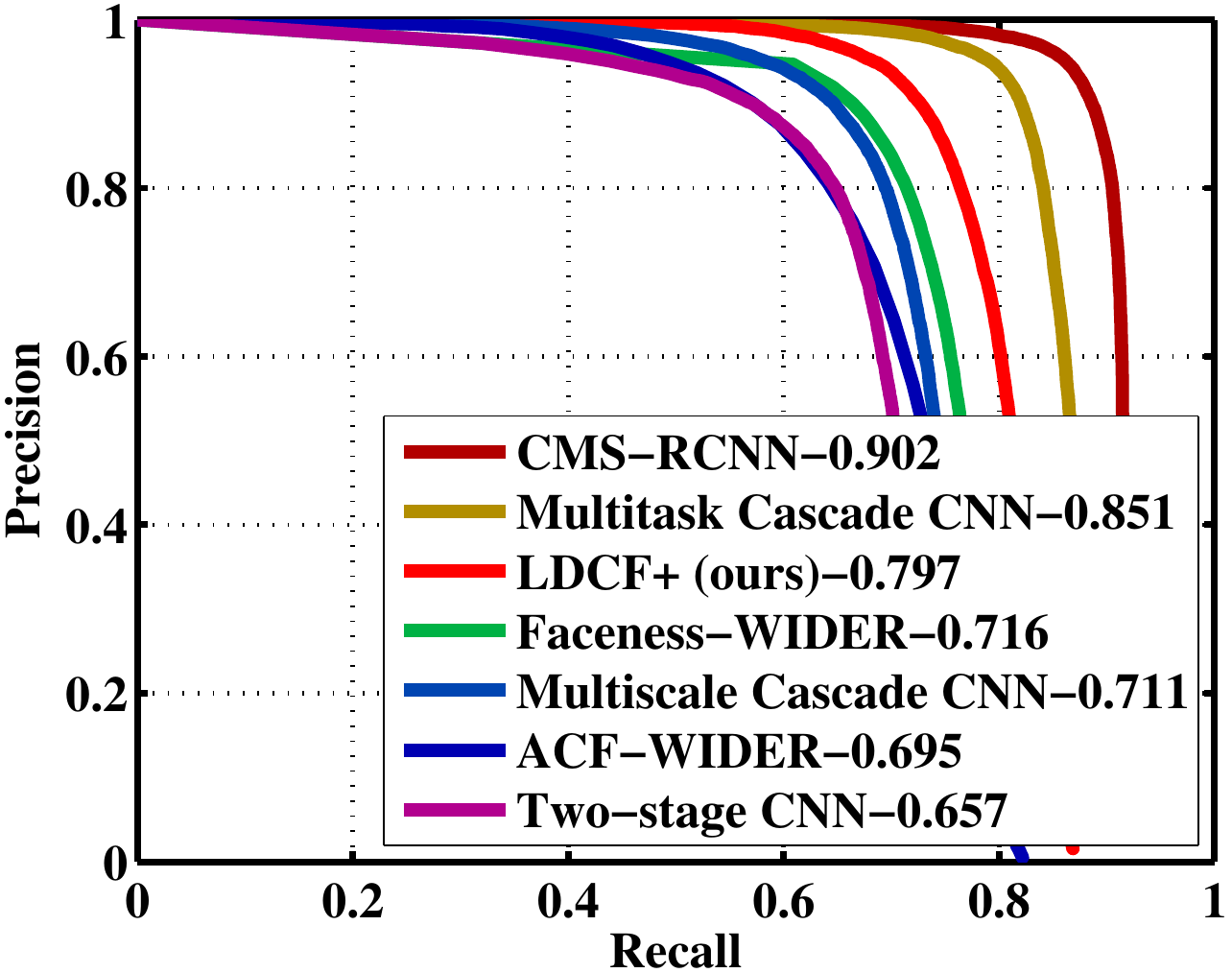} & \includegraphics[width=2in]{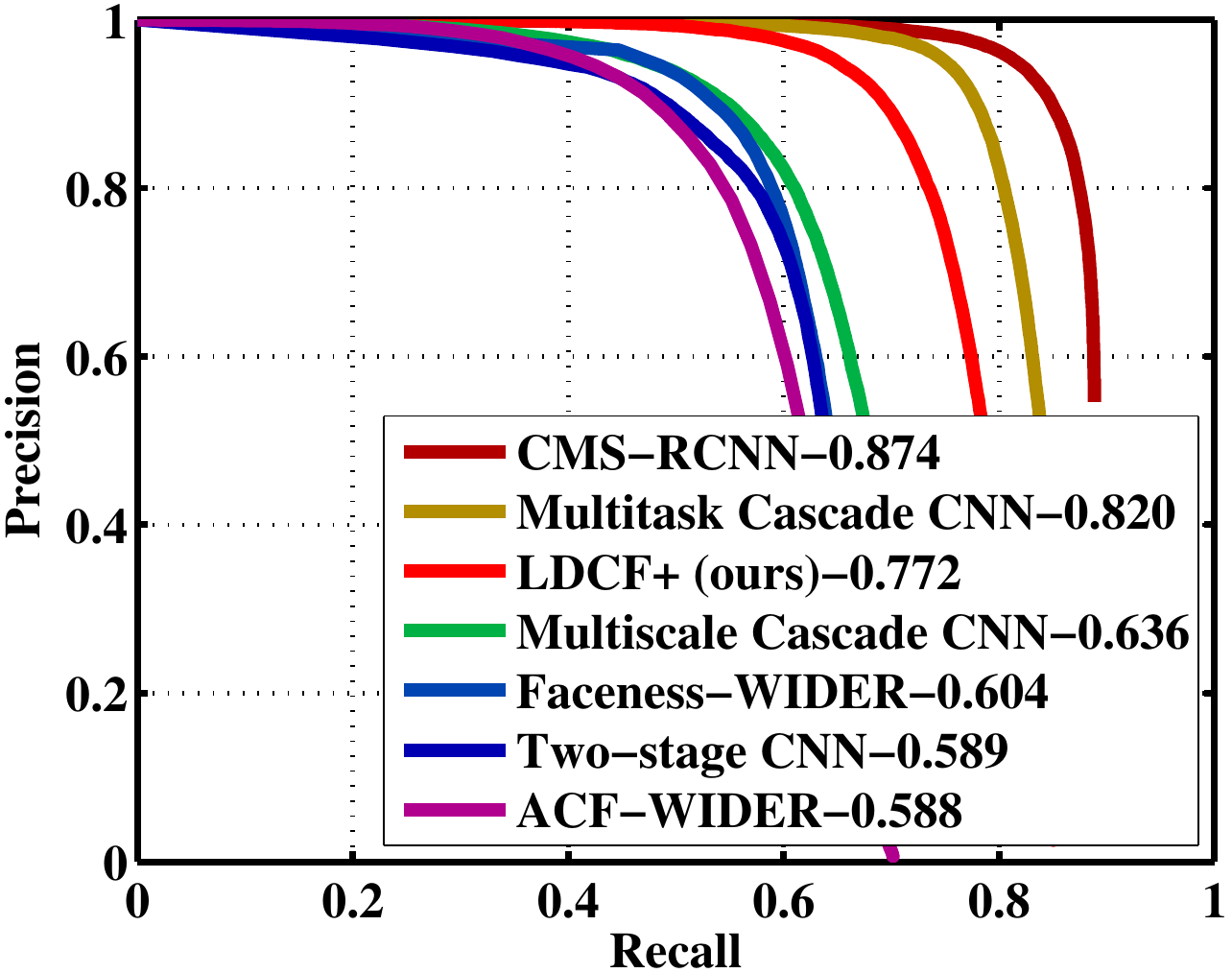} & \includegraphics[width=2in]{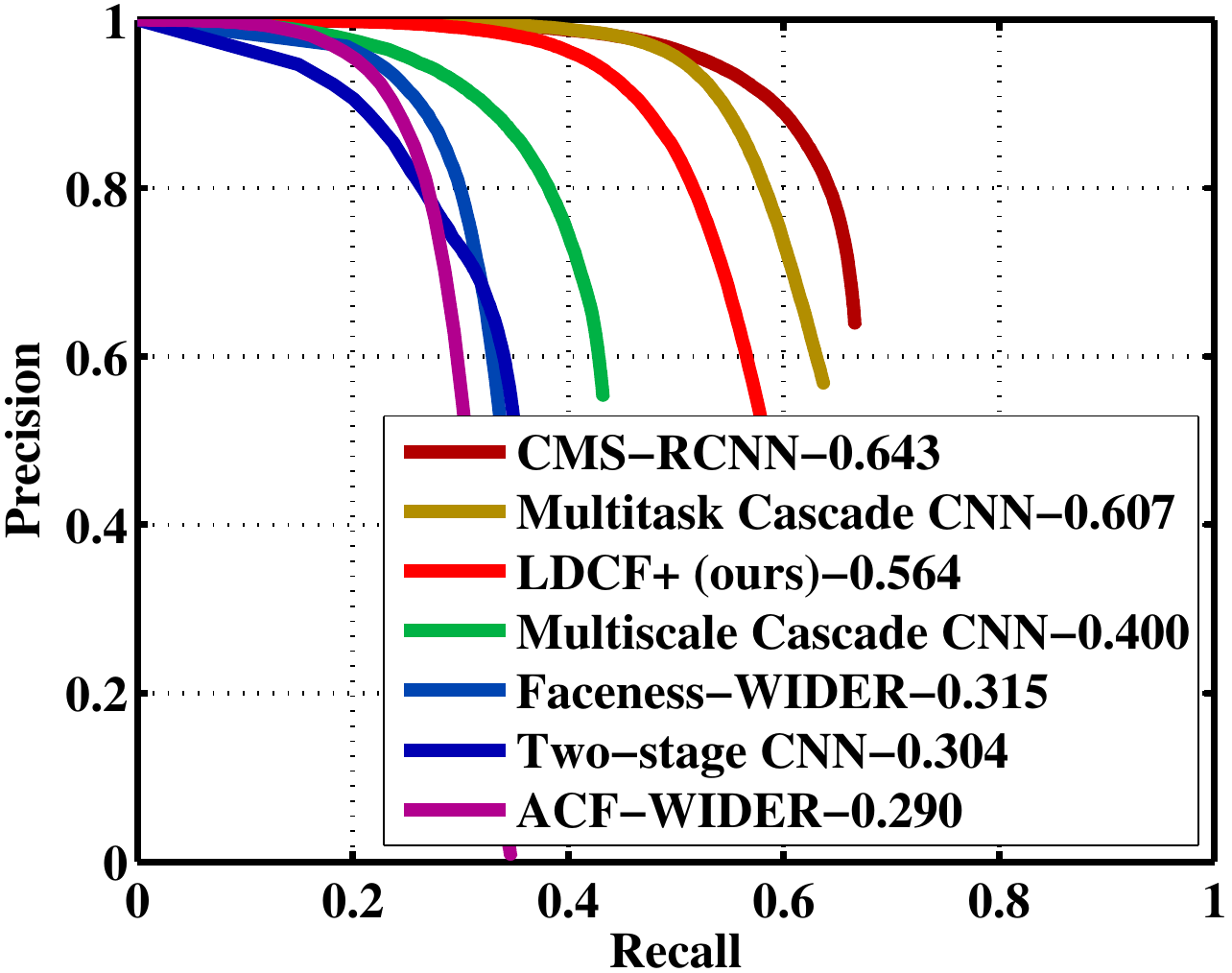}
\\
Easy set & Moderate set & Hard set
   \end{tabular}
 \caption{Results on the test set of the WIDER face dataset \cite{wider} (our curves are shown in red, LDCF+) compared with the current state-of-the-art (updated plots on January 5, 2017), Multiscale Cascade CNN \cite{wider}, Two-stage CNN \cite{wider}, ACF \cite{acfmultiview,DollarPAMI14pyramids}, Faceness \cite{faceness}, Multitask Cascade CNN \cite{DBLP:journals/corr/ZhangZL016}, and CMS-RCNN \cite{DBLP:journals/corr/ZhuZLS16}. Results are shown for the three difficulty settings described in \cite{wider}. Observe the large improvement in performance over the ACF baseline. }
 \label{fig:widertest}
\end{figure*}

\begin{figure}[!t]
\centering
\begin{tabular}{c}

\includegraphics[trim =25mm 78mm 29mm 82mm, clip=true,width=3.3in]{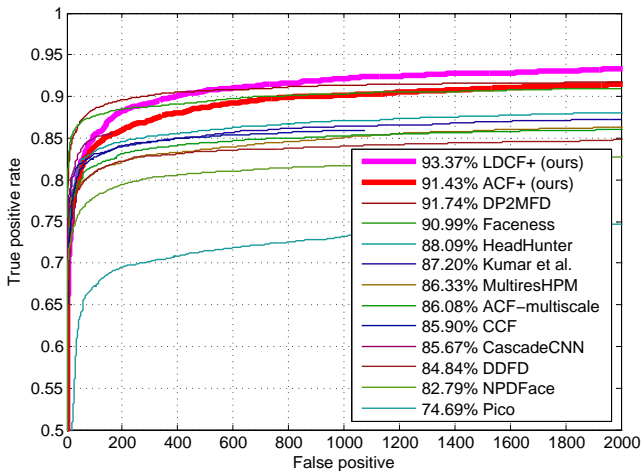}
   \end{tabular}
 \caption{Results on the FDDB \cite{fddbTech} dataset (discrete score evaluation) compared with the state-of-the-art (DP2MFD \cite{DBLP:journals/corr/RanjanPC15}, Faceness \cite{faceness}, HeadHunter \cite{bellsand}, Kumar \textit{et al.} \cite{kumar15}, MultiresHPM \cite{DBLP:journals/corr/GhiasiF15}, ACF-multiscale \cite{acfmultiview}, CCF \cite{ccf}, CascadeCNN \cite{CascadeCNN}, DDFD \cite{ddfd}, NPDFace \cite{npdface}, and Pico \cite{DBLP:journals/corr/abs-1305-4537}). True positive rate at 2000 false positives is shown for each method. Our results significantly improve over previous state-of-the-art with HOG+LUV features \cite{bellsand}, while competing with CNN approaches.}
 \label{fig:fddb}
\end{figure}

\subsection{Generalization to Face Detection}
\label{sec:faces}
The proposed training procedure is applied to face detection on the recently introduced large WIDER face dataset \cite{wider} and the commonly employed FDDB dataset \cite{fddbTech}. Specifically, we do not standardize aspect ratio and further augment the dataset as discussed in Section \ref{sec:expeval}. As shown in Table \ref{tab:fddb} on a validation set, the detection performance improve when up to depth 9 decision trees are employed. This is due to the new challenges in face detection benchmarks (e.g. rotation) not commonly found in pedestrian detection settings. Fig. \ref{fig:fddb} demonstrates consistent gains in performance, resulting in a state-of-the-art face detector (on FDDB, at 2000 false positives, 93.37\% true positive rate is achieved). On the WIDER test set, Fig. \ref{fig:widertest} depicts a large improvement over the ACF baseline results reported in \cite{wider}. The performance gap is larger on the more challenging 'moderate' and 'hard' test sets.

%
%
%
\section{Concluding Remarks}

This study presented a set of novel experiments aimed at better understanding the boosted decision tree model currently employed for many vision tasks. As CNN-based models employ extensive augmentation, we sought to investigate modifying the boosted detector to handle such augmentation as well. Careful analysis regarding the generalization power and modeling capacity, dataset size and balance, and overfitting handling produced insights as to the performance limits of such detectors, as well as state-of-the-art pedestrian and face detectors. A main takeaway of this work is in the limited representation ability of the boosted model. Unlike CNN models, which consistently benefit from additional modeling capacity (e.g. depth) and data, increasing weak learner complexity saturates early, not allowing for further gains with deeper decision trees even with extensive augmentation. Although careful data augmentation was beneficial, architectural changes for handling large datasets are required on the decision-tree training level for effective increase of modeling capacity.

\section{Acknowledgments}
We acknowledge support of associated industry partners and thank our UCSD CVRR colleagues, especially Rakesh Rajaram, for helpful discussions. We also thank the reviewers for their constructive comments.

\bibliographystyle{IEEEtran} 
\bibliography{IEEEabrv,IEEEexample}

\begin{thebibliography}{10}
\providecommand{\url}[1]{#1}
\csname url@samestyle\endcsname
\providecommand{\newblock}{\relax}
\providecommand{\bibinfo}[2]{#2}
\providecommand{\BIBentrySTDinterwordspacing}{\spaceskip=0pt\relax}
\providecommand{\BIBentryALTinterwordstretchfactor}{4}
\providecommand{\BIBentryALTinterwordspacing}{\spaceskip=\fontdimen2\font plus
\BIBentryALTinterwordstretchfactor\fontdimen3\font minus
  \fontdimen4\font\relax}
\providecommand{\BIBforeignlanguage}[2]{{%
\expandafter\ifx\csname l@#1\endcsname\relax
\typeout{** WARNING: IEEEtran.bst: No hyphenation pattern has been}%
\typeout{** loaded for the language `#1'. Using the pattern for}%
\typeout{** the default language instead.}%
\else
\language=\csname l@#1\endcsname
\fi
#2}}
\providecommand{\BIBdecl}{\relax}
\BIBdecl

\bibitem{imagenet}
A.~Krizhevsky, I.~Sutskever, and G.~E. Hinton, ``Imagenet classification with
  deep convolutional neural networks,'' in \emph{NIPS}, 2012.

\bibitem{sermanet13}
P.~Sermanet, K.~Kavukcuoglu, S.~Chintala, and Y.~LeCun, ``Pedestrian detection
  with unsupervised multistage feature learning,'' in \emph{CVPR}, 2013.

\bibitem{simonyan2014very}
K.~Simonyan and A.~Zisserman, ``Very deep convolutional networks for
  large-scale image recognition,'' \emph{ICLR}, 2014.

\bibitem{vj}
P.~Viola and M.~Jones, ``Robust real-time face detection,'' \emph{IJCV},
  vol.~57, no.~2, pp. 137--154, 2004.

\bibitem{6977320}
F.~Bartoli, G.~Lisanti, S.~Karaman, A.~D. Bagdanov, and A.~D. Bimbo,
  ``Unsupervised scene adaptation for faster multi-scale pedestrian
  detection,'' in \emph{ICPR}, 2014.

\bibitem{dollar2012pedestrian}
P.~Doll\'ar, C.~Wojek, B.~Schiele, and P.~Perona, ``Pedestrian detection: An
  evaluation of the state of the art,'' \emph{PAMI}, 2012.

\bibitem{zhang2015efficient}
S.~Zhang, C.~Bauckhage, and A.~B. Cremers, ``Efficient pedestrian detection via
  rectangular features based on a statistical shape model,'' \emph{T-ITS},
  2015.

\bibitem{de2014combinator}
F.~De~Smedt, K.~Van~Beeck, T.~Tuytelaars, and T.~Goedem{\'e}, ``The combinator:
  Optimal combination of multiple pedestrian detectors.'' in \emph{ICPR}, 2014.

\bibitem{tenyerspeds}
R.~Benenson, M.~Omran, J.~Hosang, and B.~Schiele, ``Ten years of pedestrian
  detection, what have we learned?'' in \emph{ECCV-CVRSUAD}, 2014.

\bibitem{gpuicf}
R.~Benenson, M.~Mathias, R.~Timofte, and L.~V. Gool, ``Pedestrian detection at
  100 frames per second,'' in \emph{CVPR}, 2012.

\bibitem{zhang2014center}
S.~Zhang, D.~A. Klein, C.~Bauckhage, and A.~B. Cremers, ``Center-surround
  contrast features for pedestrian detection,'' in \emph{ICPR}, 2014.

\bibitem{jones2008pedestrian}
M.~J. Jones and D.~Snow, ``Pedestrian detection using boosted features over
  many frames,'' in \emph{ICPR}, 2008.

\bibitem{icf}
P.~Doll{\'a}r, Z.~Tu, P.~Perona, and S.~Belongie, ``Integral channel
  features,'' in \emph{BMVC}, 2009.

\bibitem{acostea}
A.~D. Costea and S.~Nedevschi, ``Word channel based multiscale pedestrian
  detection without image resizing and using only one classifier,'' in
  \emph{CVPR}, 2014.

\bibitem{DollarPAMI14pyramids}
P.~Doll{\'a}r, R.~Appel, S.~Belongie, and P.~Perona, ``Fast feature pyramids
  for object detection,'' \emph{PAMI}, 2014.

\bibitem{ldcf}
W.~Nam, P.~Doll{\'a}r, and J.~H. Han, ``Local decorrelation for improved
  pedestrian detection,'' in \emph{NIPS}, 2014.

\bibitem{vehicles_TITS15}
E.~Ohn-Bar and M.~M. Trivedi, ``Learning to detect vehicles by clustering
  appearance patterns,'' \emph{T-ITS}, 2015.

\bibitem{spatialpooleecv}
S.~Paisitkriangkrai, C.~Shen, and A.~van~den Hengel, ``Strengthening the
  effectiveness of pedestrian detection with spatially pooled features,'' in
  \emph{ECCV}, 2014.

\bibitem{ccf}
B.~Yang, J.~Yan, Z.~Lei, and S.~Z. Li, ``Convolutional channel features,'' in
  \emph{ICCV}, 2015.

\bibitem{compACT}
Z.~Cai, M.~Saberian, and N.~Vasconcelos, ``Learning complexity-aware cascades
  for deep pedestrian detection,'' in \emph{ICCV}, 2015.

\bibitem{costea2015fast}
A.~D. Costea, A.~V. Vesa, and S.~Nedevschi, ``Fast pedestrian detection for
  mobile devices,'' in \emph{ITSC}, 2015.

\bibitem{boost}
R.~E. Schapire, Y.~Freund, P.~Bartlett, and W.~S. Lee, ``Boosting the margin: A
  new explanation for the effectiveness of voting methods,'' \emph{The Annals
  of Statistics}, 1998.

\bibitem{checkerboard}
S.~Zhang, R.~Benenson, and B.~Schiele, ``Filtered channel features for
  pedestrian detection,'' in \emph{CVPR}, 2015.

\bibitem{Das_ITSC14}
N.~Das, E.~Ohn-Bar, and M.~M. Trivedi, ``On performance evaluation of driver
  hand detection algorithms: Challenges, dataset, and metrics,'' in
  \emph{ITSC}, 2015.

\bibitem{hosang2015taking}
J.~Hosang, M.~Omran, R.~Benenson, and B.~Schiele, ``Taking a deeper look at
  pedestrians,'' in \emph{CVPR}, 2015.

\bibitem{rotfilt}
S.~Zhang, R.~Benenson, M.~Omran, J.~Hosang, and B.~Schiele, ``How far are we
  from solving pedestrian detection?'' in \emph{CVPR}, 2016.

\bibitem{deepparts}
Y.~Tian, P.~Luo, X.~Wang, and X.~Tang, ``Deep learning strong parts for
  pedestrian detection,'' in \emph{ICCV}, 2015.

\bibitem{bourdev2005robust}
L.~Bourdev and J.~Brandt, ``Robust object detection via soft cascade,'' in
  \emph{CVPR}, 2005.

\bibitem{bagging2}
R.~J. Quinlan, ``Bagging, boosting, and c4.5,'' in \emph{National Conference on
  Artifical Intelligence}, 1996.

\bibitem{spatialpool}
S.~Paisitkriangkrai, C.~Shen, and A.~van~den Hengel, ``Pedestrian detection
  with spatially pooled features and structured ensemble learning,''
  \emph{arXiv}, 2014.

\bibitem{quickboost}
R.~Appel, T.~Fuchs, P.~Doll{\'a}r, and P.~Perona, ``Quickly boosting decision
  trees - pruning underachieving features early,'' in \emph{ICML}, 2013.

\bibitem{6977491}
J.~Nilsson, P.~Andersson, I.~Y.~H. Gu, and J.~Fredriksson, ``Pedestrian
  detection using augmented training data,'' in \emph{ICPR}, 2014.

\bibitem{vazquez2012unsupervised}
V{\'a}zquez, David, L{\'o}pez, A.~M, and D.~Ponsa, ``Unsupervised domain
  adaptation of virtual and real worlds for pedestrian detection,'' in
  \emph{ICPR}, 2012.

\bibitem{whyandwhen}
R.~N. Rajaram, E.~Ohn-Bar, and M.~M. Trivedi, ``An exploration of why and when
  pedestrian detection fails,'' in \emph{ITSC}, 2015.

\bibitem{ohnbar_pedspatterns15}
E.~Ohn-Bar and M.~M. Trivedi, ``Can appearance patterns improve pedestrian
  detection?'' in \emph{IV}, 2015.

\bibitem{eth}
A.~Ess, B.~Leibe, K.~Schindler, and L.~V. Gool, ``A mobile vision system for
  robust multi-person tracking,'' in \emph{CVPR}, 2008.

\bibitem{zhang2015filtered}
S.~Zhang, R.~Benenson, and B.~Schiele, ``Filtered channel features for
  pedestrian detection,'' \emph{CVPR}, 2015.

\bibitem{svm}
C.-C. Chang and C.-J. Lin, ``{LIBSVM}: A library for support vector machines,''
  \emph{ACM Intell. Sys. and Tech.}, 2011.

\bibitem{multires}
D.~Park, D.~Ramanan, and C.~Fowlkes, ``Multiresolution models for object
  detection,'' in \emph{ECCV}, 2010.

\bibitem{wider}
S.~Yang, P.~Luo, C.~C. Loy, and X.~Tang, ``{WIDER FACE}: A face detection
  benchmark,'' in \emph{CVPR}, 2016.

\bibitem{fddbTech}
V.~Jain and E.~Learned-Miller, ``{FDDB}: A benchmark for face detection in
  unconstrained settings,'' University of Massachusetts, Amherst, Tech. Rep.
  UM-CS-2010-009, 2010.

\bibitem{DBLP:journals/corr/RanjanPC15}
R.~Ranjan, V.~M. Patel, and R.~Chellappa, ``A deep pyramid deformable part
  model for face detection,'' \emph{CoRR}, vol. abs/1508.04389, 2015.

\bibitem{faceness}
S.~Yang, P.~Luo, C.~C. Loy, and X.~Tang, ``From facial parts responses to face
  detection: A deep learning approach,'' in \emph{ICCV}, 2015.

\bibitem{bellsand}
M.~Mathias, R.~Benenson, M.~Pedersoli, and L.~V. Gool, ``Face detection without
  bells and whistles,'' in \emph{ECCV}, 2014.

\bibitem{kumar15}
V.~Kumar, A.~Namboodiri, and C.~V. Jawahar, ``Visual phrases for exemplar face
  detection,'' in \emph{ICCV}, 2015.

\bibitem{DBLP:journals/corr/GhiasiF15}
G.~Ghiasi and C.~C. Fowlkes, ``Occlusion coherence: Detecting and localizing
  occluded faces,'' \emph{CoRR}, vol. abs/1506.08347, 2015.

\bibitem{acfmultiview}
B.~Yang, J.~Yan, Z.~Lei, and S.~Z. Li, ``Aggregate channel features for
  multi-view face detection,'' in \emph{IJCB}, 2014.

\bibitem{CascadeCNN}
H.~Li, Z.~Lin, X.~Shen, J.~Brandt, and G.~Hua, ``A convolutional neural network
  cascade for face detection,'' in \emph{CVPR}, 2015.

\bibitem{ddfd}
S.~S. Farfade, M.~Saberian, and L.-J. Li, ``Multi-view face detection using
  deep convolutional neural networks,'' in \emph{ICMR}, 2015.

\bibitem{npdface}
S.~Liao, A.~Jain, and S.~Li, ``A fast and accurate unconstrained face
  detector,'' \emph{PAMI}, 2015.

\bibitem{DBLP:journals/corr/abs-1305-4537}
N.~Markus, M.~Frljak, I.~S. Pandzic, J.~Ahlberg, and R.~Forchheimer, ``A method
  for object detection based on pixel intensity comparisons,'' \emph{CoRR},
  vol. abs/1305.4537, 2013.

\bibitem{DBLP:journals/corr/ZhangZL016}
K.~Zhang, Z.~Zhang, Z.~Li, and Y.~Qiao, ``Joint face detection and alignment
  using multi-task cascaded convolutional networks,'' \emph{CoRR}, vol.
  abs/1604.02878, 2016.

\bibitem{DBLP:journals/corr/ZhuZLS16}
C.~Zhu, Y.~Zheng, K.~Luu, and M.~Savvides, ``{CMS-RCNN:} contextual multi-scale
  region-based {CNN} for unconstrained face detection,'' \emph{CoRR}, vol.
  abs/1606.05413, 2016.

\end{thebibliography}

\end{document}